\documentclass[sigconf]{acmart}
\usepackage{listings}
\usepackage{xcolor}
\usepackage{amsfonts}
\usepackage{amsmath}
\usepackage{graphicx}
\usepackage[bb=dsserif]{mathalpha}
\usepackage{bm}
\usepackage{float}

\usepackage{xcolor}
\usepackage{booktabs}    
\usepackage{mathtools}
\usepackage{multirow}
\usepackage{caption}
\usepackage{subcaption}
\usepackage{tabularx}
\usepackage{amsmath,bm}
\usepackage{algorithm}
\usepackage{algpseudocode}
\usepackage{wrapfig}
\usepackage{lipsum}
\usepackage{tikz}

\setcopyright{rightsretained}

\usepackage{pifont}
\definecolor{codegreen}{rgb}{0,0.6,0}
\definecolor{codegray}{rgb}{0.5,0.5,0.5}
\definecolor{codepurple}{rgb}{0.58,0,0.82}
\definecolor{backcolour}{rgb}{0.95,0.95,0.92}

\usepackage{color, colortbl}
\usepackage[first=0,last=9]{lcg}

\definecolor{LightYellow}{rgb}{0.99, 0.99, 0.59}
\definecolor{LightRed}{rgb}{0.97, 0.51, 0.47}
\definecolor{LightBlue}{rgb}{0.99, 0.59, 0.99}
\definecolor{LightGreen}{rgb}{0.59, 0.99, 0.99}

\definecolor{codegreen}{rgb}{0,0.6,0}
\definecolor{codegray}{rgb}{0.5,0.5,0.5}

\definecolor{backcolour}{RGB}{245,248,250}
\definecolor{emph}{RGB}{166,88,53}
\definecolor{nightblue}{RGB}{9,49,105}
\definecolor{keywords}{RGB}{207,33,46}
\definecolor{lightpurple}{RGB}{130,81,223}

\lstdefinestyle{mystyle}{
    backgroundcolor=\color{backcolour},   
    commentstyle=\color{codegreen},
    keywordstyle=\color{keywords},
    stringstyle=\color{nightblue},
    basicstyle=\ttfamily\footnotesize,
    breakatwhitespace=false,         
    breaklines=true,                 
    captionpos=b,                    
    keepspaces=true,                 
    showspaces=false,                
    showstringspaces=false,
    showtabs=false,                  
    tabsize=2,
    frame=shadowbox,
    emph={AutoTokenizer,AutoModelForSequenceClassification,Explainer},
    emphstyle={\color{emph}},
    emph={[2]from_pretrained,compute_table},
    emphstyle={[2]\color{lightpurple}},
    linewidth=8.0cm
}

\usepackage{amsmath}
\usepackage{empheq}
\usepackage[most]{tcolorbox}
\usepackage{enumitem}

\newtcbox{\mymath}[1][]{%
    nobeforeafter, math upper, tcbox raise base,
    enhanced, colframe=blue!30!black,
    colback=blue!10, boxrule=1pt,
    #1}
    
\newtcbox{\mymathbox}[1][]{%
    nobeforeafter, math upper, tcbox raise base,
    enhanced, colframe=red!30!black,
    colback=red!10, boxrule=1pt,
    #1}

\newtcbox{\mymathgreen}[1][]{%
    nobeforeafter, math upper, tcbox raise base,
    enhanced, colframe=green!30!black,
    colback=green!10, boxrule=1pt,
    #1}

\usepackage[capitalize,noabbrev]{cleveref}

\copyrightyear{2024} 
\acmYear{2024} 
\setcopyright{rightsretained} 

\newcommand{\EvoTF}{Evolution Transformer}

\begin{document}

\title{\EvoTF: In-Context Evolutionary Optimization}

\author{Robert Tjarko Lange}
\affiliation{%
    \institution{TU Berlin, Google DeepMind}
    \country{Germany, Japan}
 }
\email{robert.t.lange@tu-berlin.de}

\author{Yingtao Tian}
\affiliation{%
    \institution{Google DeepMind}
    \country{Japan}
 }
\email{alantian@google.com}

\author{Yujin Tang}
\affiliation{%
    \institution{Google DeepMind}
    \country{Japan}
 }
\email{yujintang@google.com}


\renewcommand{\shortauthors}{Lange et al.}

\begin{abstract}
%
Evolutionary optimization algorithms are often derived from loose biological analogies and struggle to leverage information obtained during the sequential course of optimization.
An alternative promising approach is to leverage data and directly discover powerful optimization principles via meta-optimization.
In this work, we follow such a paradigm and introduce \textit{\EvoTF}, a causal Transformer architecture, which can flexibly characterize a family of Evolution Strategies.
Given a trajectory of evaluations and search distribution statistics, \EvoTF \ outputs a performance-improving update to the search distribution. 
The architecture imposes a set of suitable inductive biases, i.e. the invariance of the distribution update to the order of population members within a generation and equivariance to the order of the search dimensions.
We train the model weights using \textit{Evolutionary Algorithm Distillation}, a technique for supervised optimization of sequence models using teacher algorithm trajectories. 
The resulting model exhibits strong in-context optimization performance and shows strong generalization capabilities to otherwise challenging neuroevolution tasks.
We analyze the resulting properties of the \EvoTF \ and propose a technique to fully self-referentially train the \EvoTF, starting from a random initialization and bootstrapping its own learning progress.
We provide an open source implementation under \url{https://github.com/RobertTLange/evosax}.
\end{abstract}

%
%

\ccsdesc[500]{Computing methodologies~Evolutionary Robotics}

\keywords{evolution strategies, machine learning}

\begin{teaserfigure}
  \centering
  \includegraphics[width=0.95\textwidth]{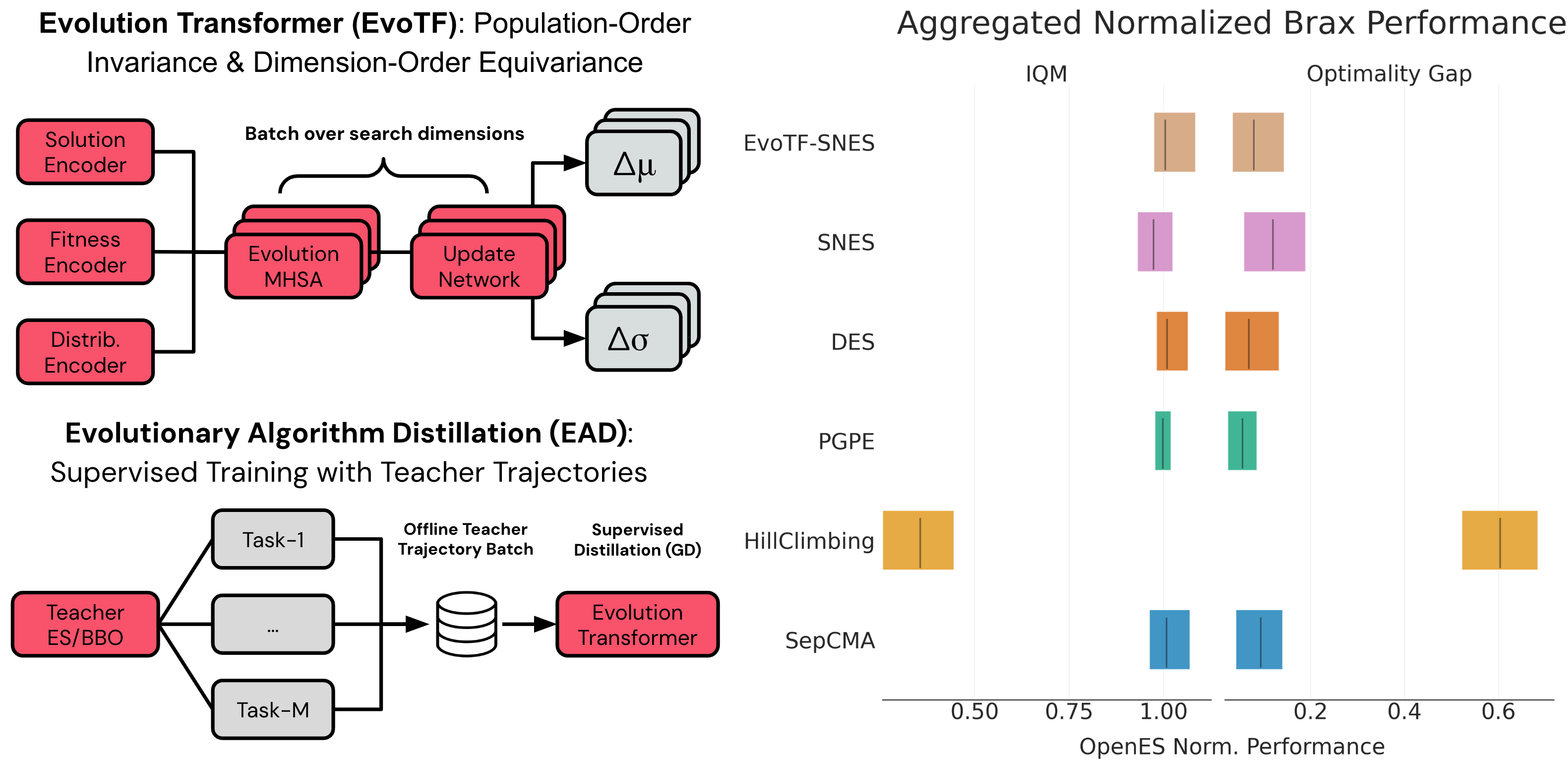}
  \caption{\textbf{Left}. \EvoTF \ overview. The model encodes three different types of information (solution space, fitness \& search distribution). The resulting per-search dimension embedding is processed by an update network using parameter sharing. The \EvoTF \ is trained by generating optimization trajectories from teacher BBO algorithms and minimizing the KL divergence between student and teacher search distribution updates. \textbf{Right}. Aggregated results on 8 Brax tasks and 5 individual runs (see SI, \cref{fig:brax_lcurves}). We report the interquartile mean and optimality gap after normalizing by the task-specific OpenAI-ES \citep{salimans2017evolution} performance. \EvoTF \ robustly outperforms Evolutionary Optimization baselines.}
  \label{fig:conceptual}
\end{teaserfigure}

\maketitle

\section{Introduction}

Evolutionary optimization \citep[EO; e.g.][]{salimans2017evolution, such2017deep} provides a scalable and simple-to-use tool for black-box optimization (BBO). Instead of having to rely on access to well-behaved gradients, such methods only require the evaluation of the function (e.g. neural network loss, training run, or physics simulator). This makes them especially well suited for otherwise challenging applications such as meta-learning or neural architecture search.
Most of the specific algorithms are derived from loose biological metaphors and are oftentimes not very adaptive to the problem at hand. Therefore, recent work has tried to address these limitations by aiming to meta-optimize attention-based modules of EO algorithms \citep{lange2023discovering_ga, lange2023discovering_es}.

Here, we extend this line of work and introduce the first fully Transformer-based Evolution Strategy (ES), which we call \textit{\EvoTF} \ (EvoTF; \cref{fig:conceptual}, Left Top).
It leverages both self- and cross-attention to implement two crucial inductive biases desirable for BBO algorithm: Invariance of the distribution update with respect to the within-generation ordering of the solutions and equivariance with respect to the ordering of the individual search dimensions.
We show that this Transformer can be successfully trained via supervised learning using optimization trajectories obtained by teacher algorithms. We name this method \textit{Evolutionary Algorithm Distillation} (EAD; \cref{fig:conceptual}, Left Bottom).
After training, the resulting model can imitate various teacher BBO algorithms and performs in-context evolutionary optimization (\cref{fig:conceptual}, Right). Furthermore, the resulting Evolution Strategy generalizes to previously unseen optimization problems, the number of search dimensions, and population members.
We propose two avenues to improve upon the teacher's performance by introducing additions to the supervised algorithm distillation training setup:
Finally, we introduce \textit{Self-Referential Evolutionary Algorithm Distillation} (SR-EAD), which can be used to self-train \EvoTF \ models: Given a parametrization, we can generate multiple perturbed trajectories, filter such, and use them to bootstrap the observed performance improvements.
Hence, it provides a promising direction for the open-ended discovery of in-context evolutionary optimizers.
Our contributions are summarized as follows:

\begin{enumerate}
    \item We introduce \textit{\EvoTF}, an architecture inducing a population-order invariant and dimension-order equivariant search update (\cref{sec:evotransformer_architecture}). After supervised training with \textit{Evolutionary Algorithm Distillation}, the Transformer has distilled various evolutionary optimizers and performs in-context learning on unseen tasks (\cref{sec:results_single_teacher}).
    \item We analyze the Transformer BBO algorithm and show empirically that it captures desirable properties such as scale-invariance and perturbation strength adaptation (\cref{sec:analysis_properties}).
    \item We compare supervised EAD with the meta-evolution of \EvoTF \ network parameters. We find that meta-evolution is feasible but tends to overfit the meta-training tasks and requires more accelerator memory (\cref{sec:meta_evolution}).
    \item We introduce \textit{Self-Referential Algorithm Distillation}, which alleviates the need for teacher algorithm specification. By perturbing the parameters of the \EvoTF, we construct a set of diverse self-generated trajectories. After performance filtering, the sequences can be used to self-referentially train the model from scratch (\cref{sec:evotransformer_adsr}).
\end{enumerate}

\begin{figure*}[h]
    \centering
    \includegraphics[width=0.95\textwidth]{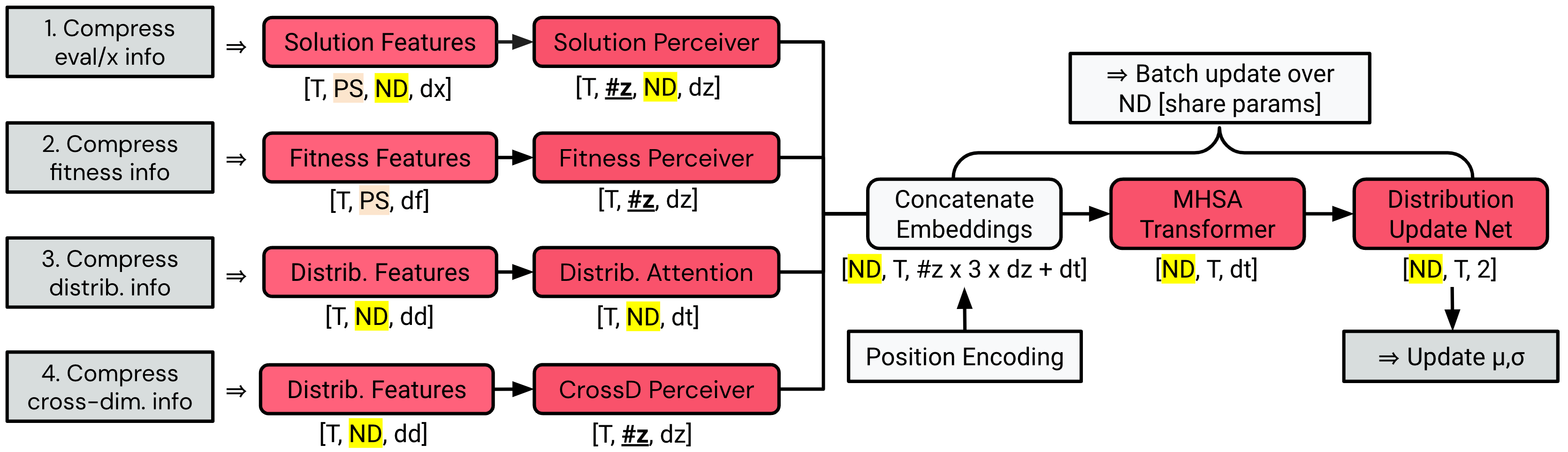}
    \caption{\EvoTF. We construct features resembling information from solution evaluations and the search distribution. They are processed by self-attention and Perceiver modules to obtain four separate embeddings. The stacked per-dimension embeddings are processed by standard Transformer encoder blocks. An MLP outputs the distribution update predictions. The model is invariant to the order of the population members and equivariant to the search dimension order.}
    \label{fig:architecture}
\end{figure*}

\newpage
\section{Related Work \& Background}

\textbf{Black-Box Optimization}. Our work develops efficient BBO algorithms with flexible Transformer-based parametrization. Given a function $f(\mathbf{x}): \mathbb{R}^D \to \mathbb{R}$ with unknown functional form, i.e. we cannot compute its derivative (or it is not well behaved), BBO seeks to find its global optimum leveraging only function evaluations:
\vspace{-0.15cm}
$$\min_\mathbf{x} f(\mathbf{x}), \text{s.t.} \ \mathbf{x}_d \in [l_d, u_d] \subset [-\infty, \infty], \forall d=1,...,D.$$
\vspace{-0.05cm}
\textbf{Evolutionary Optimization Methods}. EO provides a set of BBO algorithms inspired by the principles of biological evolution. Roughly speaking, they can be grouped into Evolution Strategies \citep{rechenberg1978evolutionsstrategien} and Genetic Algorithms (GA), which differ in the way they perform the selection and mutation of solution candidates.
Here, we focus on diagonal Gaussian ES. Given a population size $N$ and the search distribution summary statistics $\boldsymbol{\mu} \in \mathbb{R}^D, \Sigma = \boldsymbol{\sigma} \mathbf{1}_{D \times D} \in \mathbb{R}^{D \times D}$, ES sample a population of candidate solutions $X = [x_1, ..., x_N] \in \mathbb{R}^{N \times D}$ at each generation. Afterwards, the performance (or fitness) of each candidate is evaluated on the task of interest and one obtains fitness scores $F=[f_1, ..., f_N] \in \mathbb{R}^N$. The search distribution is then updated to increase the likelihood of sampling well-performing solutions, $\boldsymbol{\mu}', \boldsymbol{\sigma}' \leftarrow \texttt{UPDATE}(\boldsymbol{\mu}, \boldsymbol{\sigma}, X, F, H)$, where $H \in \mathbb{R}^{D \times D_H}$ denotes a set of ES summary statistics (e.g. momentum-like terms). There exist various types of ES including estimation-of-distribution \citep[CMA-ES,][]{hansen2006cma}, natural \citep[SNES,][]{schaul2011high, wierstra2014natural} and finite-difference-based ES \citep[OpenAI-ES,][]{salimans2017evolution}. \EvoTF \ provides a flexible parametrization of the ES-update $\texttt{UPDATE}_\theta(\boldsymbol{\mu}, \boldsymbol{\sigma}, X, F, H)$ and optimizes the set of Transformer weights $\theta$ to distill teacher BBO algorithm updates using gradient-descent.\\
\textbf{Meta-Learned Evolutionary Optimizers}. Formally, ES-update rules are inherently set of operations, i.e. the order of the population members within a generation should not affect the performed distribution change. Self-attention provides a natural inductive bias for such an operation. Previously, \citet{lange2023discovering_es, lange2023discovering_ga} constructed ES and GA algorithms, which used self- and cross-attention to process the information within a single generation. The associated parameters were then meta-evolved on a small task distribution of BBO problems. Unlike \EvoTF, the meta-evolved EO was not a sequence model, i.e. does not leverage a causal Transformer with positional encoding.\\
\textbf{Self-Referential Higher-Order Evolution}. \citet{schmidhuber1987evolutionary} first articulated the vision of self-refining Genetic Algorithms. Later on \citet{metz2021training} showed that randomly initialized black-box gradient-descent optimizers can optimize themselves using large-scale population-based training \citep{jaderberg2017population}. \citet{kirsch2022eliminating} further formulated a rule to automatically allocate resources for self-referential
meta-learning. Finally, \citep{lu2023arbitrary} provides theoretical results for infinite-order meta-evolution.\\
\textbf{Autoregressive Models for BBO}. \citet{chen2017learning} proposed to learn RNN-based BBO algorithms using privileged access to gradient computations of the fitness (Gaussian Process sampled) functions at training time. Furthermore, \citet{dery2022multi, chen2022towards} used large pre-collected datasets to train a T5 Encoder-Decoder architecture \citep{raffel2020exploring} for BBO, called OptFormer. \citet{krishnamoorthy2022generative} later on investigated the usage of generating offline data for training autoregressive BBO models. \citet{lange2024large} show that language models can be used as recombination operators for ES. In contrast, our work introduces an architecture specialized to flexibly represent ES algorithms and proposes several complementary enhancements to the supervised training pipeline to improve upon teacher algorithms. Finally, OptFormer is intended to be applied to hyperparameter optimization problems and not to train large numbers of neural network weights, unlike \EvoTF.\\
\textbf{Meta-Learned In-Context Learning \& AD for RL}. \citet{kirsch2022general, lu2023structured, team2023human} trained large sequence models using supervised and reinforcement learning without access to teacher algorithms. Instead, they only leveraged large programmatically-generated or augmented task spaces to induce in-context learning across long timescales. \citet{laskin2022context}, on the other hand, introduced the concept of Transformer-based AD in the context of Reinforcement Learning. They showed that it is possible to perform AD for actor-critic-based teacher algorithms. Afterwards, a Transformer policy shows in-context learning capabilities when evaluated on unseen environments. Our work extends these ideas to zero-order optimization and introduces a tailored architecture and AD enhancements in order to increase performance.\\
\textbf{Self-Attention \& Perceiver Cross-Attention}. The Transformer \citep{vaswani2017attention} stacks blocks of multi-head self-attention (MHSA) operations, feedforward MLP transformation, dropout and layer normalization.  At its core self-attention is a set operation which projects an input matrix $X \in \mathbb{R}^{N \times D}$ onto $D_K$-dimensional vectors $Q, K, V \in \mathbb{R}^{N \times D_K}$ called queries, keys and values, respectively:

\vspace{-0.5cm}
\begin{align*}
    \text{Attention}(X) &= \text{softmax}(QK^T/\sqrt{D_K}) V \\
    &= \text{softmax}(X W_Q (X W_K)^T/\sqrt{D_K}) X W_V.
\end{align*}
\vspace{-0.5cm}

Permuting the rows of $X$ will apply the same permutation to $\text{Attention}(X)$ \citep[e.g., ][]{lee2019set, tang2021sensory, kossen2021self}. MHSA has quadratic complexity with respect to the sequence length $N$. The Perceiver module \citep{jaegle2021perceiver} was introduced to partially alleviate this problem. Instead, it performs a cross-attention operation:

\vspace{-0.5cm}
\begin{align*}
    \text{Perceiver}(X, Z) = \text{softmax}(Z W_Q (X W_K)^T/\sqrt{D_K}) X W_V,
\end{align*}
\vspace{-0.5cm}

where $Z \in \mathbb{R}^{N_Z \times D_Z}$ denotes a set of learned latent vectors. Hence, the dimensionality of $\text{Perceiver}(X, Z) \in \mathbb{R}^{N_Z \times D_K}$ is fixed and independent of the sequence length.

\newpage
\section{\EvoTF: Population-Order In- \& Dimension- Order Equivariant Search Updates}
\label{sec:evotransformer_architecture}

We design a Transformer-based architecture that can flexibly represent an ES for different population sizes and the number of search space dimensions. To enable this endeavor, we leverage self-attention and Perceiver cross-attention to induce population-order invariance and batch over the individual search dimensions to obtain a dimension-order equivariant neural network ES.
At a high level the architecture (\cref{fig:architecture}) constructs a set of embedding features from evaluated solution candidates, their fitness scores, and search distribution statistics per generation. Such features consist of common Evolutionary Optimization ingredients such as finite difference gradients, normalized fitness scores, and evolution paths/momentum statistics (see SI for detailed input feature specification). After combining these features, we perform causal self-attention over the different generations. Finally, we output the search distribution updates per search dimension. \\
\textbf{Solution Perceiver}. Given solution candidates $X \in \mathbb{R}^{N \times D}$ sampled within a generation, we pre-compute the set of normalized features for each dimension, $\tilde{X} \in \mathbb{R}^{N \times D \times D_X}$. The information is processed by a Perceiver module applied per individual search dimension $j$. The weights are shared across the dimensions. Hence, we perform the following operations:

\vspace{-0.35cm}
\begin{align*}
    \boldsymbol{h}_d^S = \text{softmax}\left(\frac{Z W_Q (\tilde{X}[:, d] W_K)^T}{\sqrt{D_K}} \right) \tilde{X}[:, d] W_V \in \mathbb{R}^{N_Z \times D_K},
\end{align*}

for $d=1,\dots, D$. Note that this information compresses the population information into a set of $D$ tensors with a fixed shape. Thereby, the dimensionality of the resulting representation is independent of the number of population members. This together with per-dimension updates allows for general applicability to different BBO settings.\\
\textbf{Fitness Perceiver}. The fitness scores $F \in \mathbb{R}^N$ are similarly first processed into a set of features $\tilde{F} \in \mathbb{R}^{N \times D_F}$ and afterwards compressed into a set of Perceiver latent vectors, $\boldsymbol{h}^F = \mathbb{R}^{N_Z \times D_K}$.\\
\textbf{Distribution Attention \& Cross-Dimension Perceiver}. We further process a set of features capturing the search distribution dynamics (e.g. momentum and evolution paths with different time scales for both $\boldsymbol{\mu}$ and $\boldsymbol{\sigma}$). We perform both MHSA within a single dimension and Perceiver compression across dimensions.\\
\textbf{MHSA Transformer Blocks Across Time}. The four embedding steps can be computed in parallel over time, making the implementation efficient on hardware accelerators. Afterwards, we repeat and concatenate the different embeddings to construct a single stacked per-dimension representation, $\boldsymbol{h}$. The aggregated representation is further processed by a set of standard MHSA Transformer blocks with positional encoding over the timesteps/generations.\\
\textbf{Distribution Update Module}. The final representation is projected per dimension to output the proposed change in the search distribution and we perform a multiplicative update inspired by SNES \citep{schaul2011high}:

\vspace{-0.35cm}
\begin{align*}
    \boldsymbol{\mu}_{\text{EvoTF}}' &= \boldsymbol{\mu} + \eta_\mu \times \boldsymbol{\sigma} \times \texttt{EvoTransformer}_\mu(X, F, \boldsymbol{h})\\
    \boldsymbol{\sigma}_{\text{EvoTF}}' &= \boldsymbol{\sigma} \times \exp(\eta_\sigma \times \texttt{EvoTransformer}_\sigma(X, F, \boldsymbol{h})),
\end{align*}

with tune-able learning rates $\eta_\mu, \eta_\sigma$. During supervised algorithm distillation training both of them are set to 1.\\
\textbf{Evolutionary Algorithm Distillation}. We train \EvoTF \ to distill the search procedure of various teacher algorithms. More specifically, we collect batches of teacher trajectories $\{X_g, F_g, \mu_g, \sigma_g\}_{g=1}^G$ on a set of synthetic BBO tasks. We minimize the Kullback-Leibler (KL) divergence between the teacher's and proposed updated distribution at each generation with $S = \boldsymbol{\sigma}^2 \mathbf{I}_D$:

\vspace{-0.35cm}
\begin{align*}
    & \text{KL}((\boldsymbol{\mu}_{\text{EvoTF}}, \boldsymbol{S}_{\text{EvoTF}}) || (\boldsymbol{\mu}_\text{T}, \boldsymbol{S}_\text{T})) =  1/2 (\text{tr}(\boldsymbol{S}_\text{T}^{-1} \boldsymbol{S}_{\text{EvoTF}}) + \\ 
    & \log \frac{|\boldsymbol{S}_T|}{|\boldsymbol{S}_{\text{EvoTF}}|} + (\boldsymbol{\mu}_\text{T} - \boldsymbol{\mu}_{\text{EvoTF}})^T \boldsymbol{S}_\text{T}^{-1} (\boldsymbol{\mu}_\text{T} - \boldsymbol{\mu}_{\text{EvoTF}}) - D).
\end{align*}
%
\textbf{Deploying EvoTF-based Evolution Strategies}. After successful teacher distillation, we can use the trained Transformer in 'inference mode' and use it for optimization on different BBO tasks, e.g. with previously unseen numbers of search dimensions or population members. Note that during training the \EvoTF \ is never actively acting on the problems. Instead, it only has to predict the on-policy behavior of the teacher algorithm. Hence, on new evaluation problems, it has to effectively act 'off-policy' since the previous $\boldsymbol{\mu}, \boldsymbol{\sigma}$ updates were generated by the \EvoTF \ strategy itself and not by a teacher algorithm.\\
\textbf{General Implementation Details}. We use JAX \citep{jax2018github} to implement the \EvoTF \ architecture and dimension-batched search updating. We further leverage \texttt{evosax} \citep{lange2022evosax} to parallelize the generation of teacher ES algorithm sequences on hardware accelerators. The architecture uses standard Transformer encoder blocks and during training, we leverage a causal (lower diagonal) mask for autoregressively training \EvoTF \ \citep{radford2018improving} to predict the teacher algorithm's search distribution updates. Training is conducted using a single to four A100 NVIDIA GPU. For inference, we use a sliding window context of the most recent $K$ generations. This is largely due to memory constraints arising for $D>1000$ search dimensions. All \EvoTF \ ES evaluation experiments are conducted using a single V100S NVIDIA GPU. The code is published under \url{https://github.com/<anonymized_repository>}.

\newpage
\section{Supervised Evolutionary Algorithm Distillation clones various teacher BBO algorithms}



\label{sec:results_single_teacher}

\begin{figure*}[h]
    \centering
    \includegraphics[width=0.965\textwidth]{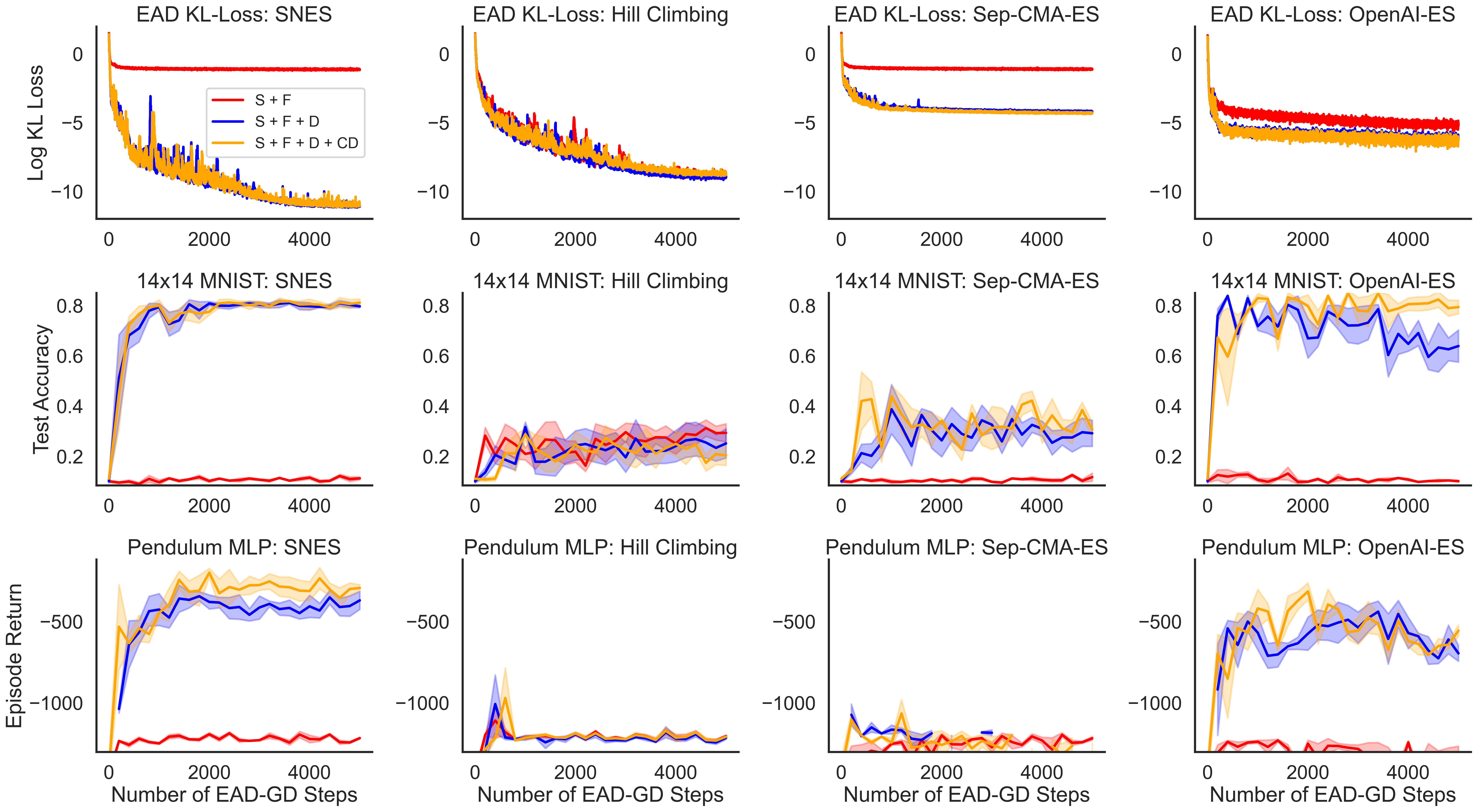}
    \caption{Evolutionary Algorithm Distillation allows EvoTF \ to distill teacher algorithms. \textbf{Top}. KL distillation loss with different Transformer modules. \textbf{Middle}. Evaluation on a 14x14 MNIST CNN Classification task throughout distillation. \textbf{Bottom}.  'S+F' uses only the Solution and Fitness Perceiver. 'S+F+D' also uses the Distribution Attention and 'S+F+D+CD' uses all network modules. Evaluation on a Pendulum MLP Control task throughout distillation. Results are averaged across 3 independent runs.}
    \label{fig:single_teacher_ad}
\end{figure*}

\textbf{Experiment Outline}. We investigate whether the proposed \EvoTF \ architecture is capable of distilling different individual BBO and ES teacher algorithms. More specifically, we focus on 4 common classes of teacher BBO algorithms: Finite-difference ES including OpenAI-ES \citep{salimans2017evolution}, estimation-of-distribution ES \citep[Sep-CMA-ES,][]{ros2008simple}, Gaussian Hill Climbing and Natural Evolution Strategies \citep{schaul2011high, wierstra2014natural}. For each individual algorithm, we generate \EvoTF \ AD training trajectories by executing it on a set of synthetic tasks (see below) with different search settings. Afterwards, we train the \EvoTF \ to auto-regressively predict the search distribution updates of the teacher algorithm. \\
\textbf{Implementation Details}. The teacher optimization trajectories are sampled from a subset of BBOB \citep{hansen2010real} benchmark functions containing different fitness landscape characteristics. For all trajectories, we use a fixed set of search dimensions ($D=5$) and population size ($N=10$). We randomize the initial mean of the search algorithm as well as its search variance. Each trajectory consists of 32 generations and we collect the solution, fitness, and distribution features. At each update step, we sample a rollout batch of 32 trajectories online before constructing the KL loss. We use pre-attention layer normalization, no dropout, and the Adam optimizer with a cosine warmup learning rate schedule. The \EvoTF \ modules use a single attention/Perceiver block and in total contain approximately 300k parameters. Throughout supervised learning, we evaluate the performance of the current network weight checkpoint with a fixed context window of $K=5$ on a set of holdout tasks including both unseen BBOB and small neuroevolution tasks.\\
\textbf{Results Interpretation}. In the first row of \cref{fig:single_teacher_ad} we show that this procedure successfully distills the considered source teacher algorithms. We note that the teacher algorithms vastly differ in their types of search update equations. Hence, the \EvoTF \ architecture design is flexible enough to represent all of these. 
Furthermore, the successful distillation of a teacher BBO algorithm requires only a handful BBOB tasks (\cref{fig:data_impact}). The performance increase does not significantly increase with the number of considered BBOB problems.
Throughout the supervised AD training procedure, we evaluate the \EvoTF \ ES on various downstream tasks. In this case, the trained \EvoTF \ model is used to perform BBO without any explicit teacher guidance. Instead, it has to perform optimization using frozen network weights and only make use of the in-context information provided by the ongoing BBO trajectory.
We find that the \EvoTF \ ES is capable of generalizing to completely unseen neuroevolution tasks such as the ant control task (\cref{fig:conceptual}). This indicates that the supervised AD procedure did not lead to overfitting, but instead has led to successful in-context evolutionary optimization.
Finally, we perform ablation studies on the modules used in the \EvoTF \ architecture (\cref{fig:single_teacher_ad}). We find that adding both the fitness and distribution information helps the model in distilling the different algorithms. The cross-dimensional Perceiver is not required. Note, that especially the distribution features are required to distill SNES, Sep-CMA-ES, and OpenAI-ES.

\newpage
\section{Analysis: \EvoTF \ Captures Desirable Evolution Strategy Properties}
\label{sec:analysis_properties}
\begin{figure*}[h]
    \centering
    \includegraphics[width=0.99\textwidth]{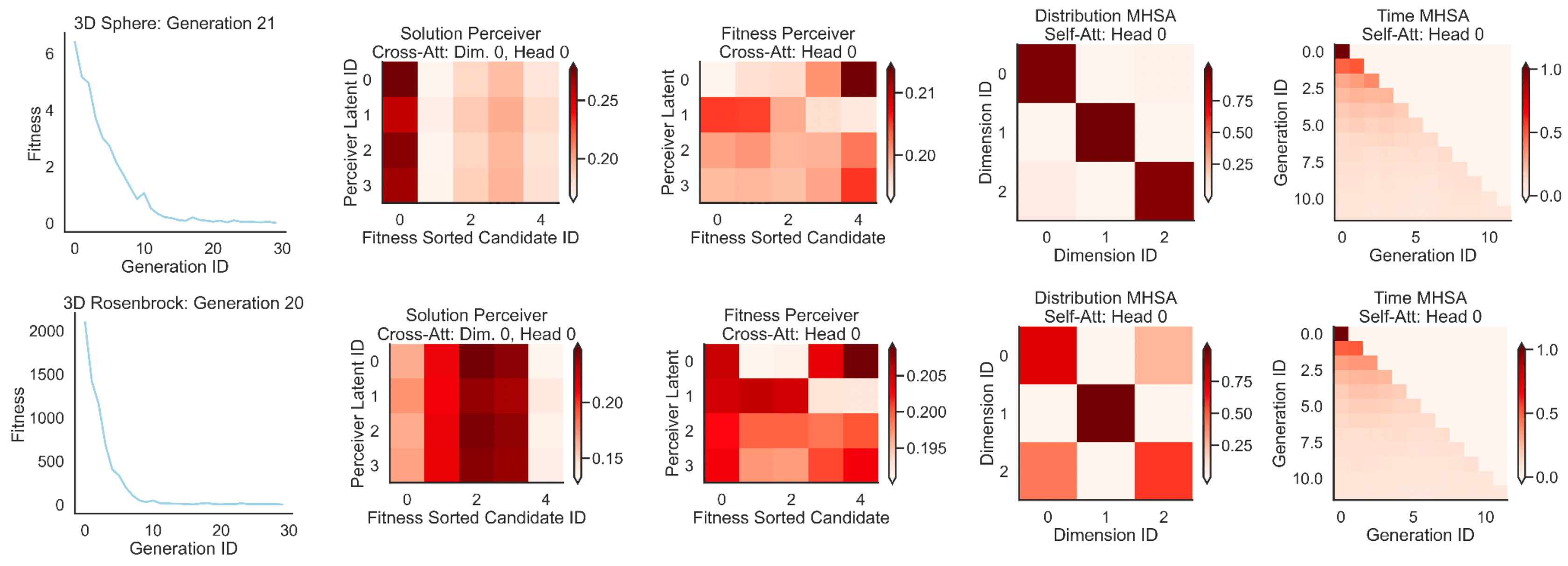}
    \caption{Self-Attention and Perceiver maps for \EvoTF \ (EAD-trained on SNES) with a single attention block at a single generation. \textbf{Top}. Separable Sphere problem. \textbf{Bottom}. Non-separable Rosenbrock problem. All problems are 3-dimensional and use 5 population members. The fitness attention assigns higher credit to the best-performing population members. The distribution attention indicates that the EvoTF correctly infers whether the fitness landscape is separable.}
    \label{fig:supp_attention}
\end{figure*}

\begin{figure}[H]
    \centering
    \includegraphics[width=0.49\textwidth]{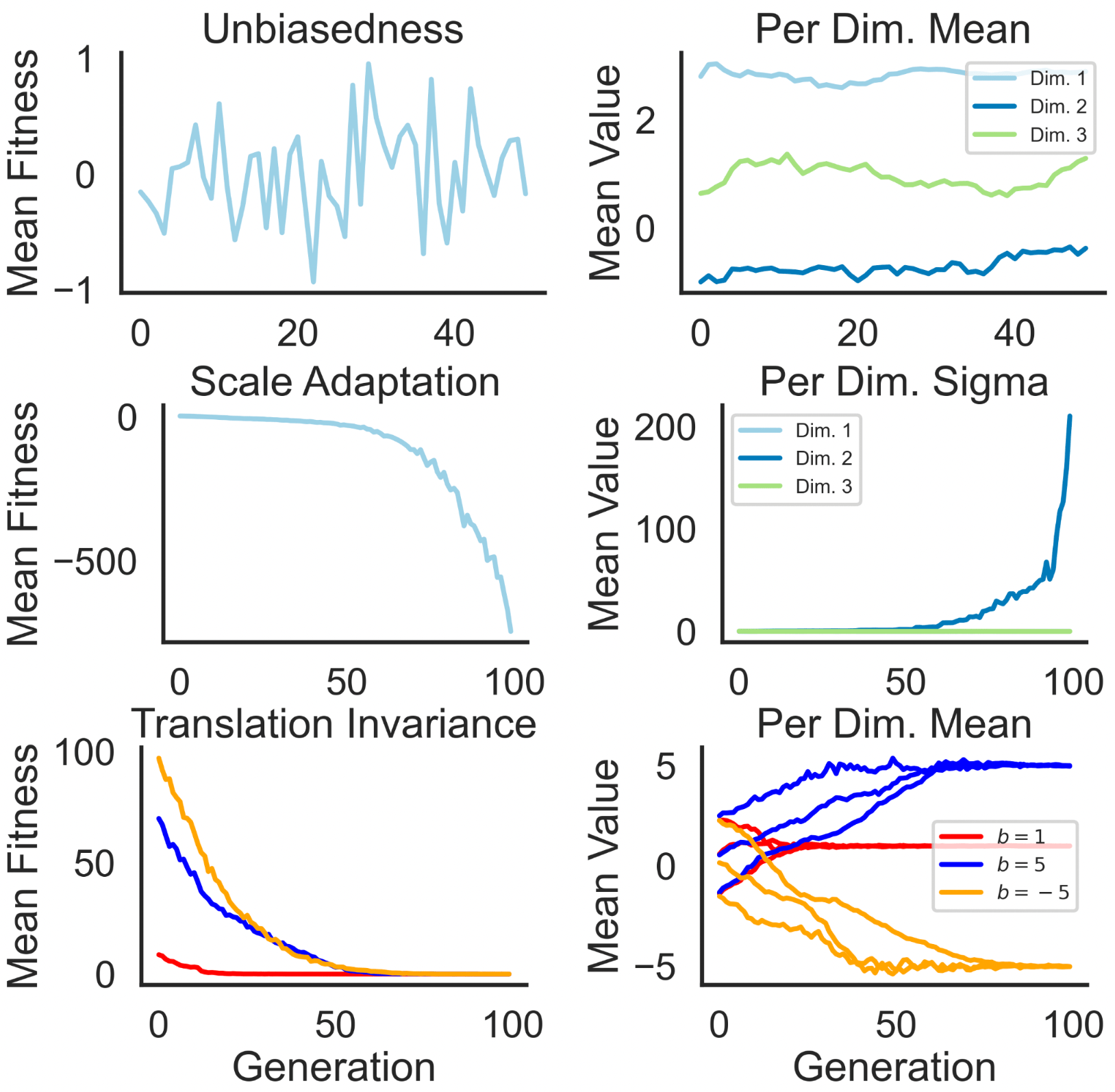}
    \caption{\EvoTF \ (SNES) properties on characteristic problems. EvoTF correctly implements the core properties of unbiasedness on the random fitness (top row), translation invariance on a Sphere Task (middle row), and scale self-adaptation on the linear fitness (bottom row). All tasks consider 3 search dimensions and 5 population members The search mean is initialized between $[-3, 3]$.}
    \label{fig:evotf_properties}
\end{figure}

\textbf{EvoTF Properties on Characteristic Problems}. After having demonstrated that EAD can successfully distill teacher BBO into the \EvoTF, we next analyze the properties of the resulting Evolution Strategy. We focus on three core properties desirable for an Evolution Strategy \citep{hansen2006cma}:
\begin{enumerate}
    \item \textbf{Unbiasedness/Stationarity}: Given a random fitness function, e.g. $f(\boldsymbol{x} \sim \mathcal{N}(0, 1)$, we do not want to observe a drift in the mean statistic, i.e. $\mathbb{E}(\boldsymbol{\mu}'| \boldsymbol{\mu}) = \boldsymbol{\mu}$.
    \item \textbf{Translation Invariance}: Given an offset to the fitness function, the performance of the ES is not going to change, e.g. $f_b(\boldsymbol{x}) = f(\boldsymbol{x} - b)$ we desire $\boldsymbol{x}_b^\star = \boldsymbol{x}^\star - b$.
    \item \textbf{Scale Self-Adaptation}: Given a linear fitness function, $f(\boldsymbol{x}) = \sum_{d=1}^D \boldsymbol{x}_d$, we desire that the perturbation strength $\boldsymbol{\sigma}$ increases over the course of the evolutionary optimization trajectory.
\end{enumerate}
We test these properties for an \EvoTF \ trained to distill the SNES teacher algorithm. After supervised EAD training, we roll out the resulting EvoTF Evolution Strategy and track the relevant statistics. In \cref{fig:evotf_properties} we show that the model correctly incorporates all the desired characteristics. The mean on a random fitness function does not show any clear bias and remains at its initialization. Furthermore, it correctly finds the optimal $\boldsymbol{x}_b^\star$ for different translation offset levels. Finally, we find that the scale quickly increases for the linear fitness function, which allows for quick adaptation and fitness improvements.\\

\textbf{\EvoTF \ Attention Maps}. Next, we visualize the attention maps for an \EvoTF \ trained to distill the SNES teacher \citep{schaul2011high}. We consider two different 3-dimensional BBO tasks: The separable Sphere task and the Rosenbrock function moderate condition number. For both cases, we consider 3 search dimensions and 5 population members.
For both problems, the fitness feature-based attention scores across the population members are highest for the best-performing members within the considered generation.
Interestingly, we observe that the distribution self-attention maps correctly identify the separability of the fitness landscape. For the Sphere task, it attributes the majority of its attention to its diagonal, indicating that each dimension-specific update mostly considers the dimension-specific evaluation information. For the non-separable Rosenbrock task, on the other hand, the attention is split across the different dimensions, hinting at a transfer of evaluation information.
The multi-head self-attention over the generation sequence, on the other hand, attributes attention across all previous generations. This indicates that the Transformer integrates information accumulated through multiple generations, providing evidence for in-context evolutionary optimization. This observation is consistent for both considered BBO problems.
Finally, the solution attention does not appear visually interpretable. This is not too surprising, given the non-linear transformations performed by the Perceiver module with the solution features.

\newpage
\section{Meta-Evolution of \EvoTF \ Weights Can Overfit The Meta-Training Task Distribution}
\label{sec:meta_evolution}

\begin{figure*}[h]
    \centering
    \includegraphics[width=0.9\textwidth]{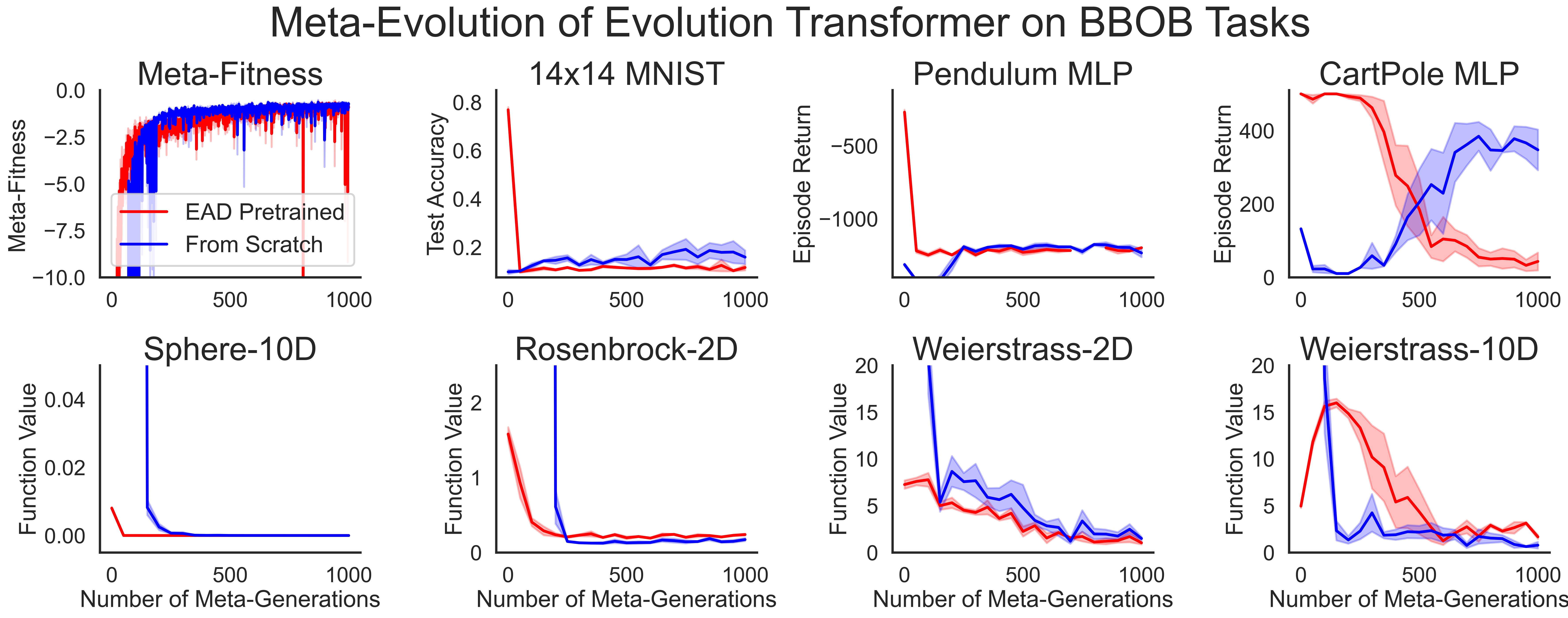}
    \caption{Meta-Evolution of \EvoTF \ weights. We train the neural network weights using meta-black-box optimization \citep{lange2023discovering_es, lange2023discovering_ga} on a set of 5 BBOB problems. We compare evolving a EvoTF parametrization from scratch with fine-tuning an EAD-pretrained SNES-EvoTF initialization. While meta-evolution quickly improves performance on BBOB tasks, it tends to not generalize to neuroevolution tasks. Results are averaged across 3 independent runs.}
    \label{fig:results_meta_evolution}
\end{figure*}

\begin{figure}[h]
    \centering
    \includegraphics[width=0.49\textwidth]{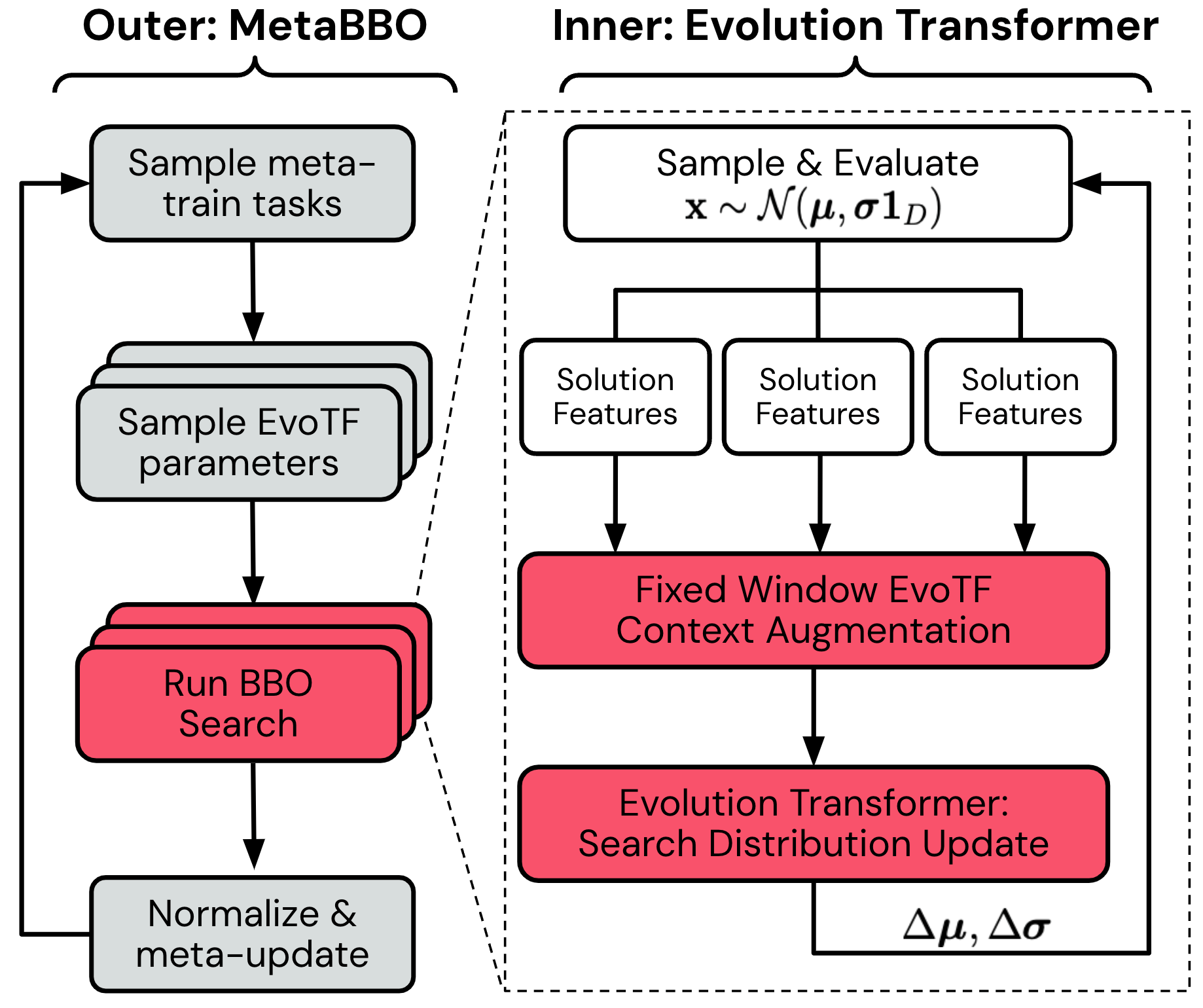}
    \caption{\EvoTF \ Meta-Black-Box Optimization (MetaBBO) procedure. We sample neural network parameters for EvoTF from a meta-optimizer. Afterwards, all EvoTransformer candidates are evaluated by running BBO on a set of tasks. We compute an aggregated meta-fitness score across the tasks and update the meta-optimizer.}
    \label{fig:meta_evolution}
\end{figure}

\textbf{Experiment Outline}. Is supervised EAD more efficient than meta-learning the weights of the \EvoTF \ directly via evolutionary optimization? Can one potentially obtain performance improvements by fine-tuning a previously distilled \EvoTF \ checkpoint via meta-evolution? To answer these questions we turn to the meta-black-box optimization (MetaBBO, \cref{fig:meta_evolution}) paradigm introduced by \citet{lange2023discovering_es, lange2023discovering_ga}. More specifically, we use a standard diagonal ES to evolve the EvoTF weights on the same set of BBOB tasks previously used for EAD. We again evaluate the performance of the resulting EvoTF checkpoint throughout meta-evolution training.\\
\textbf{Implementation Details}. We use Sep-CMA-ES \citep{ros2008simple} as the meta-ES and sample a population of 256 different EvoTF parametrizations, which we evaluate on a set of 64 BBOB tasks. Afterwards, we construct normalized meta-fitness scores for all meta-population members and update the meta-ES. This iterate meta-evolution procedure is executed for 1000 meta-generations. We consider two settings: 1) Meta-evolving EvoTF parameters starting from a random initialization and 2) starting from a previously distilled SNES teacher. Again, the entire meta-evaluation procedure is conducted using hardware acceleration and the vectorization and parallelization capabilities of JAX \citep{jax2018github, lange2022evosax}. The training is done using 4 A100 NVIDIA GPUs. Due to memory constraints, we consider a smaller EvoTF model with ca. 200k parameters and do not use the cross-dimension Perceiver module.\\
\textbf{Results Interpretation}. In \cref{fig:results_meta_evolution} we show that meta-evolving an EvoTF parametrization from a randomly initialized meta-search mean is indeed feasible and results in an ES capable of generalizing to unseen BBOB tasks and partially to neuroevolution tasks (e.g. CartPole). Only 500 meta-generations are required to obtain a competitive EvoTF-based ES.
We also investigated whether meta-evolution can act as a fine-tuning strategy after performing supervised Evolutionary Algorithm Distillation. To do so we pre-train an EvoTF to distill SNES, afterwards we initialize the meta-search mean to the previously obtained checkpoint and continue training via meta-evolution.
While such a fine-tuning exercise quickly improves performance on BBOB tasks, it dramatically decreases the performance on the neural network weight evolution evaluation tasks. This indicates a degree of overfitting to the meta-train task distribution and highlights the need for diverse meta-training coverage.
This opens up question with regards to generating a curriculum of tasks, that automatically increase the difficulty of meta-training tasks according to the capabilities of the current EvoTF.

\section{Self-Referential Evolutionary Algorithm Distillation is feasible but can be unstable}
\label{sec:evotransformer_adsr}

\begin{figure*}[h]
    \centering
    \includegraphics[width=0.9\textwidth]{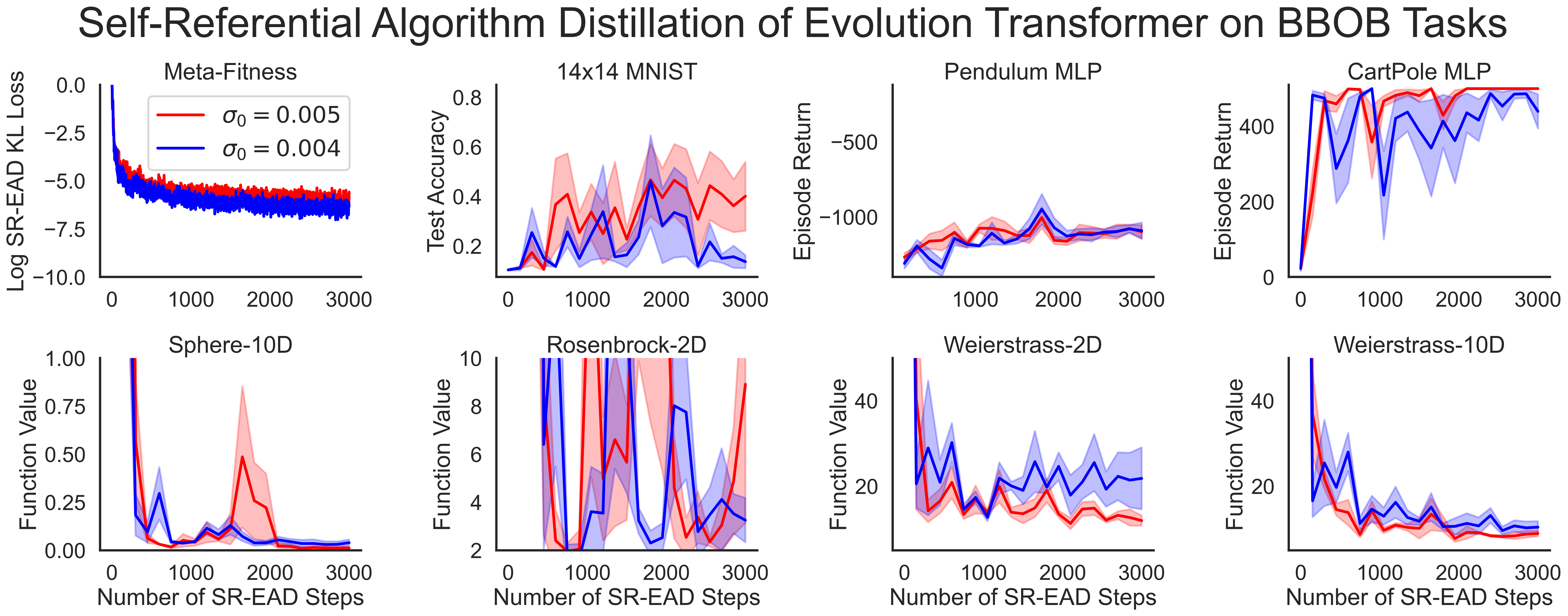}
    \caption{Self-Referential Evolutionary Algorithm Distillation of \EvoTF \ weights. By iterating the generation of offspring EvoTF parametrizations, the generation and filtering of BBO trajectories, and self-distillation of such filtered trajectories, we obtain a self-referentially trained evolution strategy. Results are averaged across 3 independent runs.}
    \label{fig:results_meta_sr_ead}
\end{figure*}

\begin{figure}[h]
    \centering
    \includegraphics[width=0.475\textwidth]{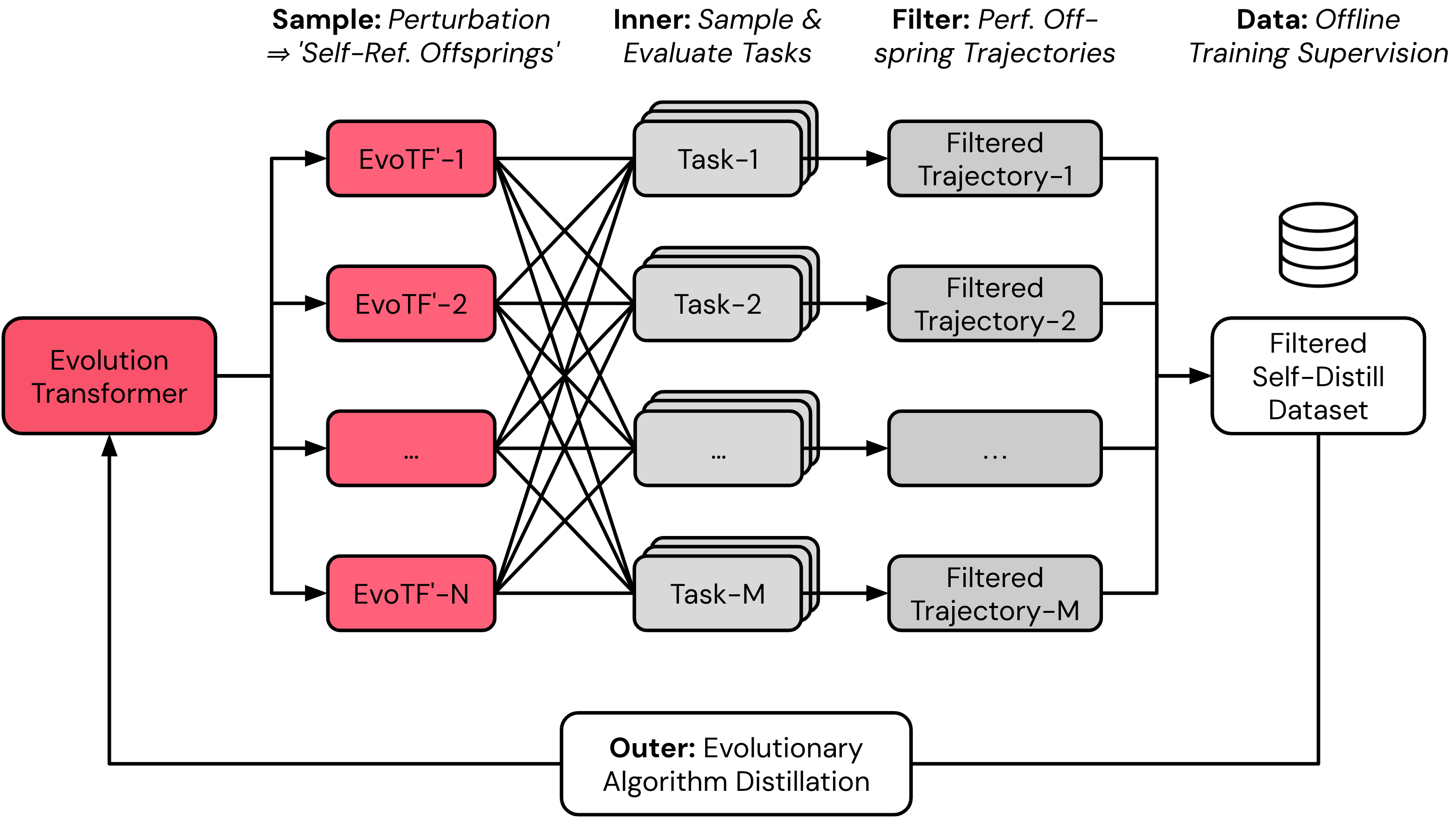}
    \caption{Self-Referential Evolutionary Algorithm Distillation Procedure for \EvoTF \ without the need for a teacher algorithm or meta-optimizer.}
    \label{fig:meta_sr_ead}
\end{figure}

\textbf{Experiment Outline}. We wondered whether it is possible to completely alleviate the need for a teacher algorithm (in EAD) or a meta-black-box optimization algorithm (e.g. Sep-CMA-ES). More specifically, we propose to leverage random perturbations of an EvoTF checkpoint to generate different so-called 'self-referential offspring' (\cref{fig:meta_sr_ead}). Afterwards, these offspring generate diverse optimization trajectories on a fixed set of tasks. We then filter the collected trajectories by their performance. Only the best trajectories are used to train the EvoTF with algorithm distillation. This procedure iterates between perturbation, trajectory generation, performance filtering and distillation. Thereby, at each iteration of the procedure, the EvoTF from the previous iteration can bootstrap its behavior based on previously observed performance improvements of its perturbed versions.\\
\textbf{Implementation Details}. As a proof of concept, we use simple scalar Gaussian perturbations to the current EvoTF weight checkpoint to generate $N=64$ EvoTF offspring. We compare two different perturbation strengths used to generate offspring, $\sigma_0 \in \{0.004, 0.005\}$. We very slowly (exponentially) decay the strength throughout training. At each iteration, we collect the performance of the offspring on the previously used subset of 5 BBOB tasks. All of the offspring are evaluated on the same randomly sampled tasks and we select the best BBO trajectory observed by all offspring to be added to the distillation batch. We proceed by performing a single gradient descent distillation update and iterating the procedure. We conduct training on 4 A100 NVIDIA GPUs leveraging JAX-parallelization of the trajectory generation step.\\
\textbf{Results Interpretation}. In \cref{fig:results_meta_sr_ead} we show that is indeed feasible to perform self-referential learning without the need for an explicit improvement algorithm in the form of a meta-optimizer or teacher algorithm. The performance on the BBOB tasks quickly improves above random. Furthermore, on a subset of neuroevolution tasks such as CartPole and MNIST classification, the SR-EAD-based EvoTF can perform the optimization of MLP weights. On the Pendulum control task, on the other hand, self-referential training appears unstable. After an initial improvement, the performance drops again. This observation appears robust to the choice of perturbation strength.

\newpage
\section{Conclusion}

\textbf{Summary}. We introduced the \EvoTF, a sequence model fully based on the Transformer architecture that characterizes a family of ES. Using Evolutionary Algorithm Distillation, we cloned various BBO algorithms, which can afterwards be deployed on unseen optimization tasks. It provides a novel point of view for data-driven evolutionary optimization. While using meta-evolution to directly optimize the Transformer parameters is feasible, it results in overfitting to the considered meta-training problems. Finally, we introduced Self-Referential Evolutionary Algorithm Distillation, which alleviates the need for a teacher or meta-optimization algorithm. Instead, it uses random parameter perturbations to bootstrap observed self-improvements. We hypothesize that this procedure can facilitate the open-ended discovery of novel in-context evolutionary optimization algorithms.\\
\textbf{Limitations}. While our Transformer-based model can be trained efficiently on modern hardware accelerators, the deployment comes with large memory requirements, especially for a large number of search dimensions or population members. Therefore, we had to employ a sliding context window. This ultimately limits the power of in-context learning. While we provide empirical insights into the behavior of the Evolution Transformer, we still lack a full mechanistic understanding of such a 'black-box BBO'. Finally, self-referential training can be unstable and appears to jump between local optima. Stabilization will require a better theoretical understanding of the learning dynamics induced by mixing evolutionary perturbation/data generation and gradient descent-based self-distillation.\\
\textbf{Future Work}. We are interested in the open-ended discovery of novel algorithms leveraging SR-EAD. While we experimented solely with evolutionary optimization in the inner loop, the framework is more general and could be applied to any sequential algorithm with a suitable filtering mechanism. Furthermore, recent developments in state space models may provide an opportunity to address the context length limitations. Finally, further benchmarking on a more diverse task sets will be required \citep{lange2024neuroevobench}.

\newpage

\bibliographystyle{ACM-Reference-Format}
\renewcommand*{\bibfont}{\tiny}
\bibliography{bibliography}


\appendix
\newpage
.
\begin{figure*}[h]
    \centering
    \includegraphics[width=0.99\textwidth]{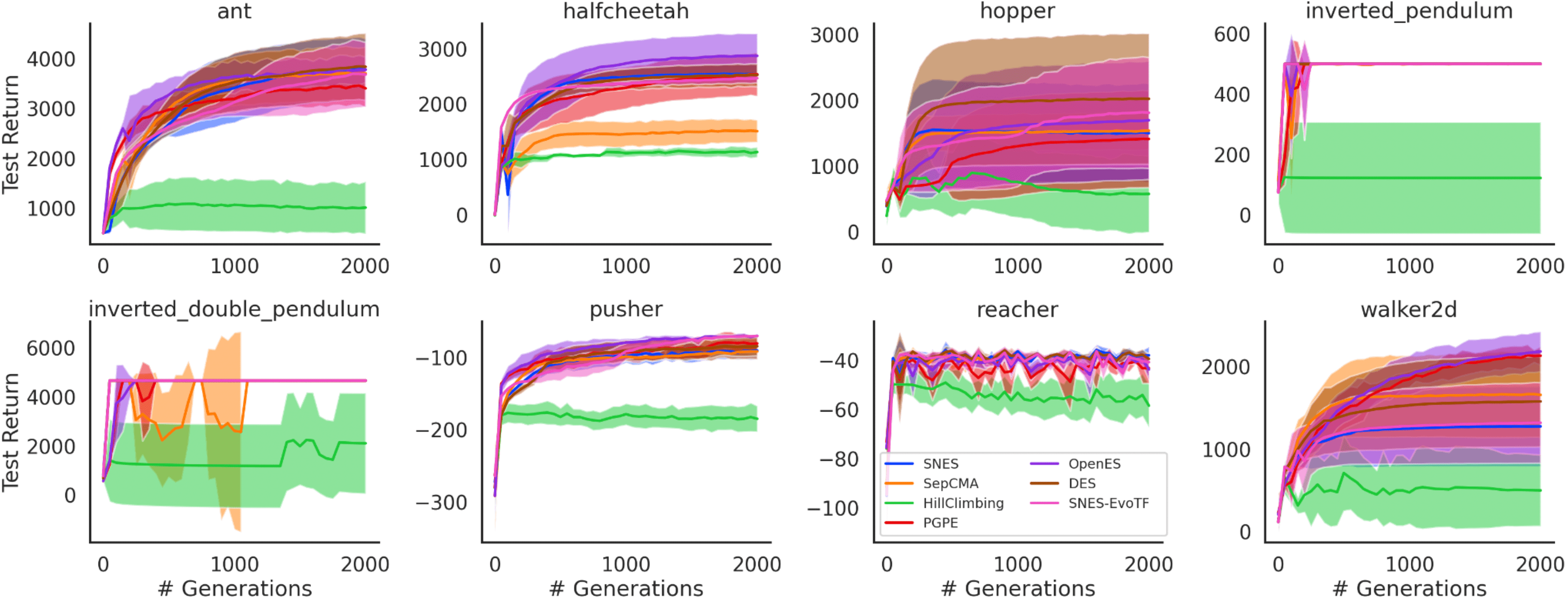}
    \caption{Full Learning Curve on Brax Tasks using a population size of $N=128$ and a Tanh Control Network. A SNES-distilled EvoTF performs very competitively. Results are averaged across 5 independent runs.}
    \label{fig:brax_lcurves}
\end{figure*}
\newpage

\section{Additional Results}
\subsection{Brax Learning Curves}

We use the Brax \citep{freeman2021brax} set of continuous control tasks and the EvoJAX \citep{tang2022evojax} evaluation wrapper. We use the standard 4-layer tanh activation network to control various robots. The population size is $N=128$ and we tune all baselines and EvoTF hyperparameters using a small grid search over $\sigma_0$ and the learning rates. See \cref{fig:brax_lcurves}.

\subsection{Impact of Task Distribution for EAD}

\begin{figure*}[h]
    \centering
    \includegraphics[width=0.99\textwidth]{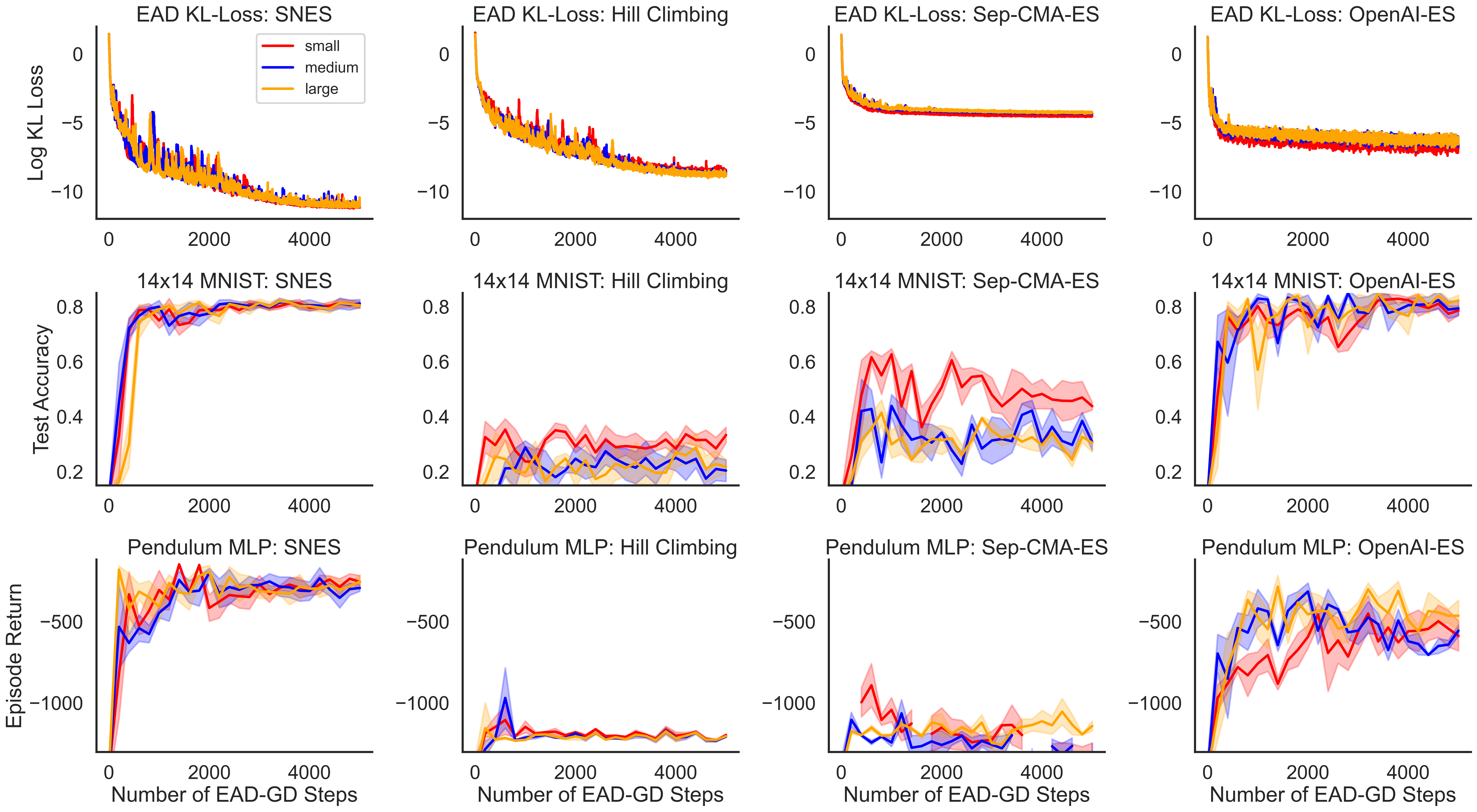}
    \caption{Impact of task distribution for EAD performance. We compare using 1 (small), 5 (medium) and 10 (large) BBOB functions with random offsets and mean intialization. Successful distillation does not require many optimization tasks. Results are averaged across 5 independent runs.}
    \label{fig:data_impact}
\end{figure*}

We compare the impact of the task distribution used to generate distillation trajectories. See \cref{fig:data_impact}.

\begin{table}[H]
\centering
\footnotesize
\begin{tabular}{ |p{1.65cm}|p{2.4cm}|p{2.25cm}|p{1cm}|}
 \hline
 Function & Reference & Property & Tasks\\
 \hline
 Sphere   &  \citet[p.\ 5, ][]{hansen2010real} & Separable (Indep.) & Small\\
 \hline
 Rosenbrock   &  \citet[p.\ 40, ][]{hansen2010real}   & Moderate Condition & Medium\\
 Discus & \citet[p.\ 55, ][]{hansen2010real} & High Condition & Medium\\
 Rastrigin   &  \citet[p.\ 75, ][]{hansen2010real}   & Multi-Modal (Local) & Medium\\
 Schwefel & \citet[p.\ 100, ][]{hansen2010real} & Multi-Modal (Global) & Medium\\
 \hline
 BuecheRastrigin & \citet[p.\ 20, ][]{hansen2010real} & Separable (Indep.) & Large \\
 AttractiveSector & \citet[p.\ 30, ][]{hansen2010real} & Moderate Condition & Large\\
 Weierstrass & \citet[p.\ 80, ][]{hansen2010real} & Multi-Modal (Global) & Large\\
 SchaffersF7 & \citet[p.\ 85, ][]{hansen2010real} & Multi-Modal (Global) & Large\\
 GriewankRosen & \citet[p.\ 95, ][]{hansen2010real} & Multi-Modal (Global) & Large\\
 \hline
\end{tabular}
\caption{Meta-BBO BBOB-Based Task Families.}
\label{table:bbob}
\end{table}

Throughout all main experiments, we use the 'Medium' set of BBOB tasks.

\section{Evolution Transformer Features \& Model Hyperparameters}

\begin{table}[H]
\centering
\small
\begin{tabular}{|r|l||r|l|}
\hline \hline
Parameter & Value & Parameter & Value \\ 
\hline
Number Transformer Blocks                & 1   & Number of MHSA Heads        & 2      \\
Number of Perceiver Latents                & 4   & Perceiver Latent Dim        & 32      \\
Embedding Dimension              & 64   &    Batchsize    &   32    \\
Sequence Length              & 32   &    Learning rate    &   0.0015    \\
Gradient Clip Norm              & 1   &    Train steps    &   5000    \\

\hline \hline
\end{tabular}
\caption{Evolution Transformer Hyperparameters.}
\end{table}

The features used as inputs include:

\begin{enumerate}
    \item Solution: Normalized difference from mean, squared normalized difference from mean, difference from best seen solution, normalized difference from best seen solution, etc.
    \item Fitness: Improvement boolean, z-score, centered rank, normalized in range, SNES weights, etc.
    \item Distribution: Finite-difference gradient, SNES gradient, evolution paths, momentum paths
\end{enumerate}

\section{Meta-Evolution Hyperparameters}

\begin{table}[H]
\centering
\small
\begin{tabular}{|r|l||r|l|}
\hline \hline
Parameter & Value & Parameter & Value \\ 
\hline
Population Size                & 256   & \# Generations        & 1000      \\
MetaBBO                & Sep-CMA-ES   & Perturbation Strength        & 0.005      \\
Task Batch Size                & 64   &     &  \\

\hline \hline
\end{tabular}
\caption{Meta-evolution hyperparameters.}
\end{table}

\section{Self-Referential Evolutionary Algorithm Distillation (SR-EAD) Hyperparameters}

\begin{table}[H]
\centering
\small
\begin{tabular}{|r|l||r|l|}
\hline \hline
Parameter & Value & Parameter & Value \\ 
\hline
Population Size                & 64   & Perturbation Strength        & 0.004      \\
Exponential Decay                & 0.99999   & Filtering        & Best      \\
\hline \hline
\end{tabular}
\caption{SR-EAD hyperparameters.}
\end{table}

\section{Software Dependencies}

The codebase will be open-sourced under Apache 2.0 license and publicly available under \url{https://github.com/<anonymous>}.  All training loops and ES are implemented in JAX \citep{jax2018github}.
All visualizations were done using Matplotlib \citep[]{hunter_2007} and Seaborn \citep[BSD-3-Clause License]{waskom_2021}. Finally, the numerical analysis was supported by NumPy \citep[BSD-3-Clause License]{harris2020array}. Furthermore, we used the following libraries: Evosax: \citet{lange2022evosax}, Gymnax: \citet{gymnax2022github}, Evojax: \citet{tang2022evojax}, Brax: \citet{freeman2021brax}.

\section{Compute Requirements \& Experiment Organization}

The experiments were organized using the \texttt{MLE-Infrastructure} \citep[MIT license]{mle_infrastructure2021github} training management system. 

Simulations were conducted on a high-performance cluster using between 3 and 5 independent runs (random seeds). We mainly rely on individual V100S and up to 4 A100 NVIDIA GPUs. The experiments take between 2 and 5 hours wall clock time.

\end{document}